\documentclass[conference]{IEEEtran}
\IEEEoverridecommandlockouts
\usepackage{cite}
\usepackage{amsmath,amssymb,amsfonts}
\usepackage{algorithmic}
\usepackage{graphicx}
\usepackage{textcomp}
\usepackage{xcolor}
\def\BibTeX{{\rm B\kern-.05em{\sc i\kern-.025em b}\kern-.08em
    T\kern-.1667em\lower.7ex\hbox{E}\kern-.125emX}}

\newcommand{\changed}[1]{#1}
\newcommand{\matt}[1]{\textcolor{black}{#1}} 
    
\begin{document}

\title{Longitudinal Robot Learning from Demonstration with Care Providers in a Home Environment}

\author{Nina Moorman*, Julianna Schalkwyk, Vriksha Srihari, Qingyu Xiao, Kamel Alrashedy,\\ Hongseok Jeong, Kiersten Lange, Matthew B. Luebbers, Matthew Gombolay
\thanks{*Corresponding author: nmoorman3@gatech.edu}%
\thanks{Authors affiliated with Georgia Institute of Technology, Atlanta, GA, USA.}%
\thanks{This research was supported by NSF grants IIS-2340177 and IIS-2112633.}%
}


\maketitle

\begin{abstract}
Learning from demonstration (LfD) methods enable non-expert end users to teach robots novel skills without explicit programming. 
However most evaluations of the usability of LfD with non-experts has been conducted in controlled laboratory environments with a robotics experimenter present. In this work we identify non-expert end users' key barriers when teaching robots via demonstration without live robotics expert feedback \changed{in a home environment.}
\changed{In our human subjects experiment we support the non-expert end users through two forms of demonstrator guidance developed in prior work: pre-training and adaptive feedback.}
Towards the ecological validity of the evaluation, we conduct this experimentation \changed{over multiple visits}, with a population of care providers.
\changed{Finally, we propose to open source the resulting LfD dataset of care providers teaching a robot assistive tasks over multiple visits to a home environment.}
\end{abstract}

\begin{IEEEkeywords}
learning from demonstration, translational research, assistive robotics
\end{IEEEkeywords}

\section{Introduction}
\changed{Household robots will need to interact with and adapt to unstructured, dynamic human environments, presenting a challenge for robot policies trained in controlled, laboratory-like settings \cite{olatunji2026robots}.}
While increasing the scale of training data could help address this challenge,
collecting such real-world data is expensive, time consuming, \changed{and risks exposing private information}. 
Alternatively, in-home supplemental learning from end users would enable personalization of \matt{robot} behavior to \changed{those} users' needs and \changed{environments}, without requiring the same prohibitive scale of outside training data. 
Learning from Demonstration (LfD) methods enable \matt{robots} to learn skills from user demonstrations. Various prior works have developed instructional materials \cite{moorman2023investigating, cakmak2014teaching} and \matt{feedback mechanisms} \cite{moorman2026teaching, srivastava2023generating} to enable non-experts to teach robots via LfD. 

This work seeks to \matt{ascertain} the remaining challenges non-expert users will face when teaching robots via LfD.
As prior work has shown that demonstrators improve at providing kinesthetic demonstrations over the course of multiple sessions \cite{aliasghari2024non}, we propose a multi-visit, multi-task evaluation.
To improve the ecological validity of our evaluation, we conduct our study in a realistic home environment where non-expert end users will define \matt{and demonstrate} their own \matt{custom} \changed{unimanual and bimanual manipulation} tasks. 
We \matt{will} recruit participants with no prior robotics or computer science experience and, over the course of three visits, provide them with automated training and feedback, to determine, with such support, what challenges they face when teaching a robot assistive household tasks.

\changed{In this work we propose to contribute the following:}
\begin{enumerate}
    \item \changed{We design a human subjects experiment to characterize current challenges and opportunities for deploying LfD systems with non-expert end users provided with pre-training and adaptive feedback, in a home environment.} 
    \item \changed{We will open source the subsequent multi-visit dataset of non-experts teaching robots a series of assistive tasks in a home environment.} 
\end{enumerate}


\section{Demonstration System}


To detect task-relevant objects and produce segmentation masks, we process an RGB-D stream from an Intel RealSense camera with YOLOE \cite{wang2025yoloe}. The corresponding masked depth measurements are back-projected into object-specific 3D point clouds. To estimate object pose, point clouds are registered against predefined object shapes using Iterative Closest Point (ICP) \cite{besl1992method}, producing a 6D pose estimate for tracked objects. 

\matt{For robot skill learning, we} use Cartesian Probabilistic Motion Primitives (ProMPs) \cite{paraschos2013probabilistic} over end effector and tracked object poses. We choose to employ ProMPs as they can learn from limited demonstrations, can be trained quickly, and are robust to changing object poses \changed{allowing for generalization to different environment setups,
compared to Behavior Cloning approaches which often struggle to generalize \cite{codevilla2019exploring} or Vision-Language Action models that may struggle to adapt to people's specific preferences without time consuming finetuning.}

We develop an interactive interface for \matt{providing} kinesthetic demonstrations. 
The interface allows participants to plan \matt{a set of low-level} skills \matt{to} teach to accomplish tasks in \matt{a given} domain (\matt{called a} task decomposition), provide demonstrations for each skill, and create recipes of skills to accomplish \changed{tasks}. 
The interface additionally supports the participant through \changed{pre-training and through adaptive feedback}. 

\underline{\changed{Pre-Training (PT)}}

We provide participants with \changed{PT developed in prior work \cite{moorman2023investigating} where,} for \matt{a set of pre-defined domains,} the participant first attempts to break down a task in that domain, then is shown \changed{videos of an} optimal task decomposition and how \changed{those} skills are taught by a robotics expert via video recordings\matt{.} 
PT allows the participant to improve upon their task decompositions \textit{on novel domains} without the need for live feedback from an expert demonstrator \cite{moorman2023investigating}.

\newpage
\underline{\changed{Adaptive Feedback (AF)}}
\paragraph*{Foundation Model (FM) Feedback}
\matt{FM feedback} \changed{provides guidance on} whether \matt{a} participant's planned task decomposition can adequately generalize to downstream tasks \cite{moorman2026teaching}. As input, the FM feedback \matt{module} takes in descriptions of the domain, environment, and known tasks. 
To support novice users in defining and describing \matt{novel} domains \matt{and} tasks, we develop an additional FM feedback module, \matt{beyond that of prior work, which} iteratively asks clarification questions until the \matt{domain} description has the necessary amount of detail for the \changed{aforementioned} task decomposition feedback to be helpful.



\paragraph*{Real Robot Replay (RRR) Feedback} The robot rolls out the learned policy, allowing the user to directly observe its performance \matt{in-situ} \cite{moorman2026teaching}.

\paragraph*{Augmented Reality (AR) Feedback} The robot rolls out the learned policy in AR \changed{to serve as a visual debugging tool \cite{luebbers2021arc}}, allowing the user to directly observe \changed{policy} performance \matt{and generalizability} without time consuming or dangerous environmental interactions\cite{moorman2026teaching}.


\section{\matt{Data Collection \& Human Evaluation}} 

We conduct \changed{our evaluation} in the \matt{Georgia Tech} Aware Home \cite{kidd1999aware}, an authentic home environment. We recruit from a population of local \changed{care providers of older adults, both formal (doctors, nurses, physical and occupational therapists), and informal  \cite{cantor1979neighbors} (family, friends, or neighbors who interface with the older adult on at least a monthly basis).}



\paragraph*{Procedure}
\begin{figure}
    \centering
    \includegraphics[width=\linewidth]{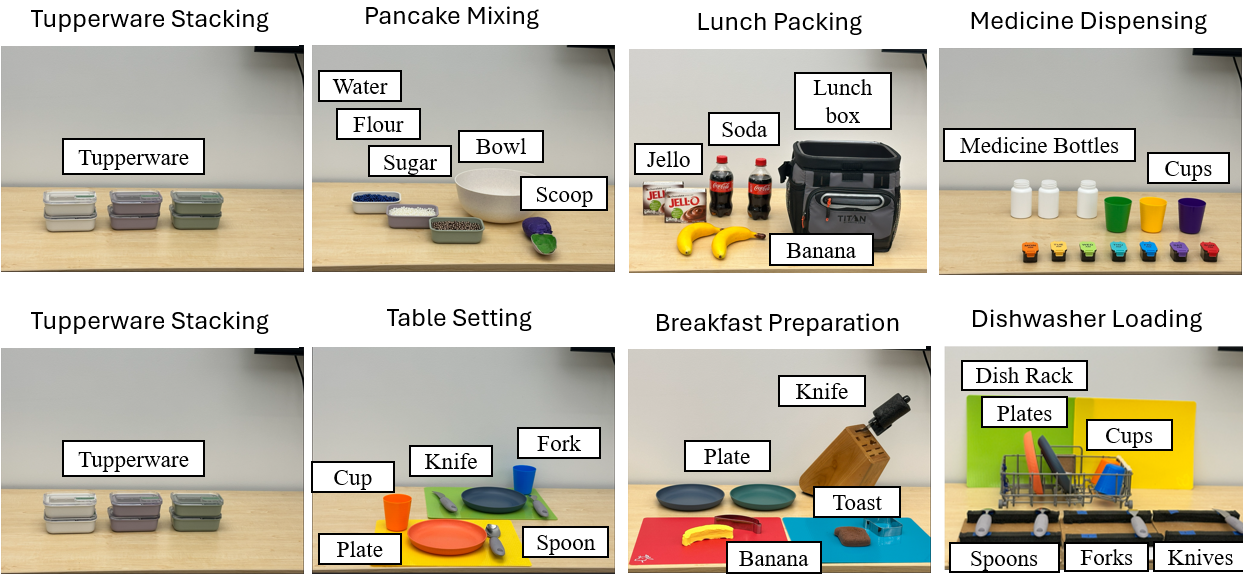}
    \caption{Domains for visit 1 (first row) and visit 2 (second row). 
    \matt{Tupperware stacking serves as a kinesthetic on-boarding task for both visits.} 
    \changed{These domains were chosen to represent a range of activities of daily living. Additionally, we balance task complexity and distribution of unimanual to bimanual tasks between visits.}}
    \label{fig:domains}
\end{figure}
In this work, participants teach tasks to \matt{a pair of} JACO 2 arms over the course of three visits. In the first visit, participants experience the \changed{PT} \cite{moorman2023investigating} where they acquire experience teaching the robot a series of tasks \matt{via LfD} in various domains (depicted in the first row of Figure \ref{fig:domains}). In the second visit, participants teach the robot \matt{additional tasks} in various domains (depicted in the second row of Figure \ref{fig:domains}) with  \changed{all three forms of AF available (FM, AR, and RRR feedback)}. In the third and final visit, participants \changed{are tasked to define} their own \matt{custom} domains and tasks, \changed{with AF available}.

\paragraph*{Conditions and Research Questions (RQ)}
Our \changed{between-subjects} experiment has two conditions. In the \changed{\textbf{PT+AF}} condition, participants obtain \changed{PT} in visit 1, then experience \changed{AF} feedback for visit 2 and visit 3. In the \textbf{\changed{AF}} condition, participants experience \changed{AF} feedback for visit 2 and visit 3, but do not participate in visit 1. Our RQs include:

\begin{itemize}
    \item \textbf{RQ1:} \textit{Are there differences between the \changed{PT+AF} condition and the \changed{AF} condition in terms of task performance?} After each visit, we evaluate task completion percentage \changed{using a pre-defined rubric}.
    
    \item \textbf{RQ2:} \textit{Are there differences between the \changed{PT+AF} condition and the \changed{AF} condition in terms of alignment between predicted and actual robot performance?} \changed{Alignment is calculated as the absolute difference between task performance and normalized perceived performance \cite{moorman2026teaching}.}
    
    
    \item \textbf{RQ3:} \textit{Are there differences between the \changed{PT+AF} condition and the \changed{AF} condition in terms of user experience?} The user experience metrics of interest include usability and acceptance \cite{belanche2012integrating} and learned trust \cite{jian2000foundations}, and workload \cite{hart1988development}. 
\end{itemize}
\changed{Statistical comparisons between conditions will employ multiple linear regressions with mixed effects,
controlling for time-on-task, number of demonstrations, and visit count.}

\section{Outcomes}
\begin{figure}
    \centering
    \includegraphics[width=\linewidth]{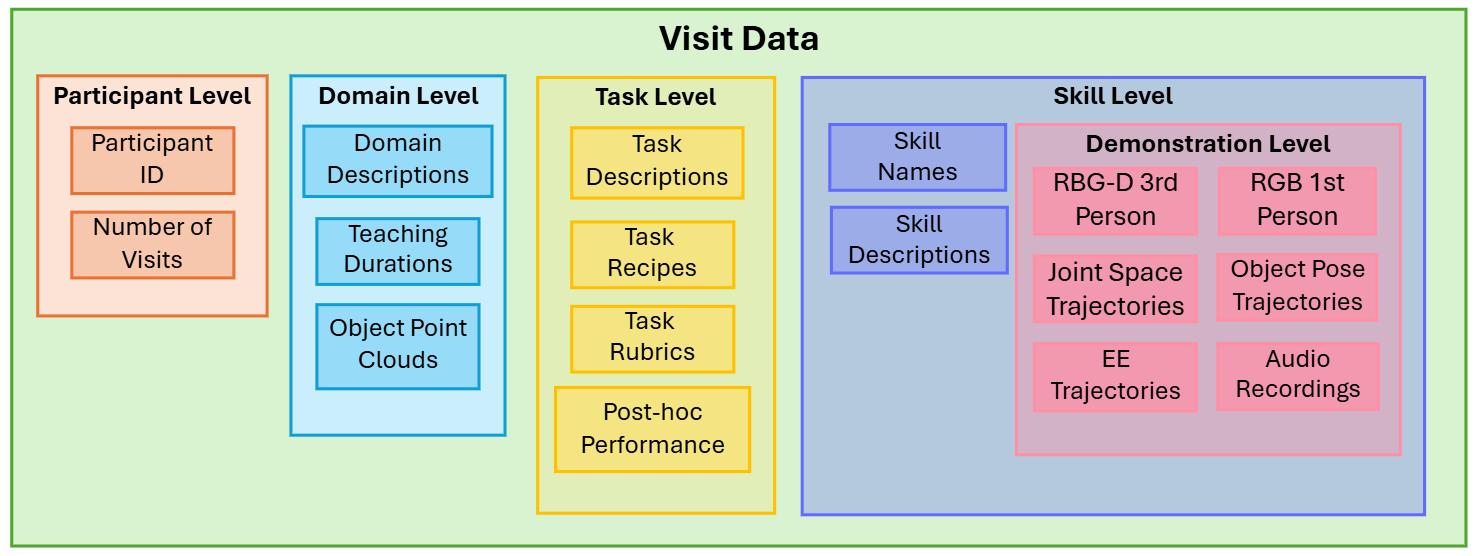}
    \caption{This figure depicts the dataset structure \changed{where each datapoint represents one visit, with information about the participant, domains, tasks, and demonstrated skills.}}
    \label{fig:dataset}
\end{figure}

The outcomes from this work will be twofold. 
\changed{We will open source the data collected, supplementing existing datasets of demonstrated robot skills such as MIME \cite{sharma2018multiple} (which focuses on unimanual, multi-step tasks), or RoboPro \cite{roboPro2026} (focusing on bimanual, short-horizon tasks) with realistic assistive tasks trained in-situ over multiple visits by care providers.}
\matt{This} will benefit the field by enabling standardized benchmarking \matt{for} interactive robot learning algorithmic development and evaluation. The dataset structure is depicted in Figure \ref{fig:dataset}. 

\changed{For each of the eleven domains (or eight domains in the AF condition), participants employ their skill library to accomplish six tasks. Natural language descriptions of domains, tasks, and skills will be included in the dataset, along with the point-clouds of all objects in each domain. Demonstration data will include end-effector and joint trajectories, participant audio, and egocentric and exocentric video footage.}


Furthermore, in addition to reporting upon the human-factors findings related to our three \changed{research questions}, we plan to report on lessons learned and propose design guidelines for researchers \matt{and practitioners} deploying LfD systems in home environments \matt{for assistive tasks}, \changed{designing for} care providers, \matt{and collecting multi-visit demonstration datasets}. 

\newpage
\bibliographystyle{IEEEtran}
\bibliography{refs}

\end{document}